\documentclass[letterpaper, 10 pt, conference]{ieeeconf}  

\IEEEoverridecommandlockouts                              

\usepackage{url}
\usepackage{graphics} 
\usepackage{graphicx} 
\usepackage{mathptmx} 
\usepackage{times} 
\usepackage{amsmath} 
\usepackage{amssymb}  
\usepackage{color}
\usepackage{xcolor}
\usepackage{subfig}
\usepackage{epstopdf}
\usepackage{balance}
\usepackage{wrapfig}
\usepackage{adjustbox}
\usepackage{booktabs, makecell}
\usepackage{multirow}
\usepackage{times}
\usepackage{textcomp}
\usepackage{floatrow}
\usepackage{algpseudocode}
\usepackage{dsfont}
\usepackage{array}
\usepackage{caption}
\usepackage[ruled,linesnumbered]{algorithm2e}
\usepackage[font=small,labelfont=bf]{caption}
\usepackage{multicol}
\DeclareMathAlphabet{\mathcal}{OMS}{cmsy}{m}{n}

\title{\LARGE \bf
Semantic Mapping with Simultaneous Object Detection and Localization
}

\author{Zhen Zeng \hspace{0.5cm} Yunwen Zhou \hspace{0.5cm} Odest Chadwicke Jenkins\hspace{0.5cm} Karthik Desingh  
\thanks{Z. Zeng, Y. Zhou, O.C. Jenkins, K.  Desingh are with the Department of Electrical Engineering and Computer Science, University of Michigan, Ann Arbor, MI, USA, 48109-2121. {\tt\small [zengzhen|zhezhou|zsui|ocj]@umich.edu}}
}

\begin{document}
\twocolumn[{
\renewcommand\twocolumn[1][]{#1}%
\maketitle
\vspace*{-0.8\baselineskip}
\begin{center}
    \centering
    \includegraphics[width=\textwidth]{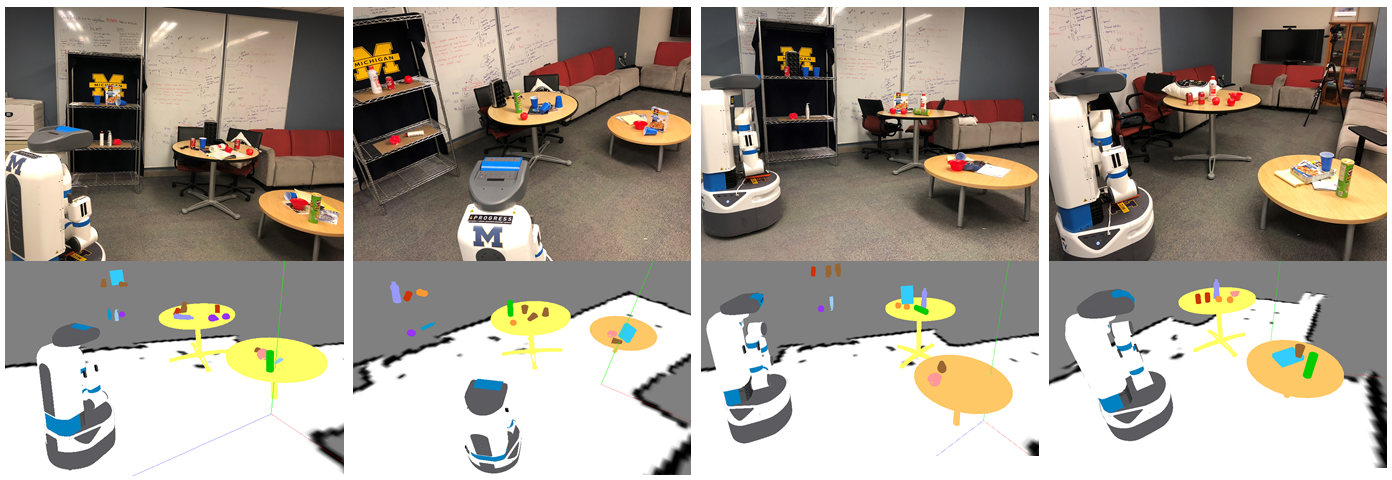}
    \captionof{figure}{Robot semantically maps a student lounge in four different visits. Each column shows an RGB snapshot of the environment, together with the corresponding semantic map composed by the detected and localized objects. We propose Contextual Temporal Mapping ({\em CT-Map}) method to simultaneously detect objects and localize their 6 DOF pose given streaming RGB-D observations. To achieve this, we probabilistically formulate semantic mapping problem as a problem of belief estimation over object classes and poses. We use Conditional Random Field (CRF) to model contextual relations between objects and temporal consistency of object poses. (Best viewed in color)}
    \label{fig:teaser}
\end{center}%
\vspace*{0.8\baselineskip}
}]
{\renewcommand\thefootnote{\fnsymbol{footnote}}
 \footnotetext{Z. Zeng, Y. Zhou, O.C. Jenkins, K.  Desingh are with the Department of Electrical Engineering and Computer Science, University of Michigan, Ann Arbor, MI, USA, 48109-2121. {\tt\small [zengzhen|ywchow|ocj|kdesingh]@umich.edu}}
}

%

\begin{abstract}
We present a filtering-based method for semantic mapping to simultaneously detect objects and localize their 6 degree-of-freedom pose. For our method, called Contextual Temporal Mapping (or {\em CT-Map}), we represent the semantic map as a belief over object classes and poses across an observed scene. Inference for the semantic mapping problem is then modeled in the form of a Conditional Random Field (CRF). {\em CT-Map} is a CRF that considers two forms of relationship potentials to account for contextual relations between objects and temporal consistency of object poses, as well as a measurement potential on observations. A particle filtering algorithm is then proposed to perform inference in the {\em CT-Map} model. We demonstrate the efficacy of the {\em CT-Map} method with a Michigan Progress Fetch robot equipped with a RGB-D sensor. Our results demonstrate that the particle filtering based inference of {\em CT-Map} provides improved object detection and pose estimation with respect to baseline methods that treat observations as independent samples of a scene.
\end{abstract}


\section{Introduction}

For robots to effectively operate and interact with objects, they need to understand not only the metric geometry of their surroundings but also its semantic aspects. When requested to organize a room or search for an object, robots must be able to reason about object locations and plan goal-directed mobile manipulation accordingly. We aim to enable robots to semantically map the world at the object level, where the representation of the world is a belief over object classes and poses. With the recent advances in object detection via neural networks, we have stronger building blocks for semantic mapping. Yet, such object detections are often times noisy in the wild, due to biases and insufficient diversity in training dataset. In our work, we aim to be robust to false detections from such networks. We model the object class as part of our hidden state for generative inference, rather than making hard decisions on class labels as given by the detector.

Given streaming RGB-D observations, our goal is to infer object classes and poses that explain observations, while accounting for \textbf{contextual} relations between objects and \textbf{temporal} consistency of object poses. Instead of assuming that every object is independent in the environment, we aim to explicitly model the \textit{object-object} contextual relations during semantic mapping. More specifically, objects from the same category (e.g., food category) are expected to co-occur more often than objects that belong to different categories. Additionally, physical plausibility should be enforced to prevent objects from intersecting with each other, as well as floating in the air.



Temporal consistency of object poses also plays an important role in semantic mapping. Objects could stay where they were observed in the past, or gradually change their semantic locations over time. 
Under cases of occlusion, modeling temporal consistency can potentially help the localization of partially observed objects. Through temporal consistency modeling, the robot could gain a notion of object permanence, i.e., believing that objects continue to exist even when they are not being directly observed.

Considering both contextual and temporal factors in semantic mapping, we propose the \textbf{Contextual Temporal Mapping} (\textbf{{\em CT-MAP}}) method to simultaneously infer object classes and poses. Examples of semantic maps generated by {\em CT-Map} are shown in Figure \ref{fig:teaser}. To avoid deterministically representing the world as a collection of recognized objects with poses, we maintain a belief over the object classes and poses across observations.

For generative inference, {\em CT-MAP} probabilistically formalizes the semantic mapping problem in the form of a Conditional Random Field (CRF). Dependencies in the CRF model capture the following aspects: 1) compatibility between the latent semantic mapping variables and observations, 2) contextual relations between objects,  and 3) temporal consistency of object poses. 
We propose a particle filtering based algorithm to perform generative inference in {\em CT-MAP}, inspired by Limketkai et al~\cite{limketkai2007crf}.

We evaluate the proposed semantic mapping method {\em CT-MAP} with the Michigan Progress Fetch robot. The performance of {\em CT-MAP} is quantitatively evaluated in terms of object detection and pose estimation accuracy. We show that {\em CT-MAP} is effective in simultaneously detecting and localizing objects in cluttered scenes. We demonstrate object detection performance superior to Faster R-CNN~\cite{ren2017faster}, and accurate 6 DOF object pose estimation compared to 3D registration methods such as ICP, and FPFH~\cite{rusu2009fast}. We also highlight examples in which our method benefits from modeling temporal consistency of object poses and object contextual relations. 

\section{Related Work}
Our work semantically maps the world through simultaneous object detection and 6 DOF object pose estimation. Contextual relations between objects and temporal consistency of object poses are being modeled for better scene understanding. Here we discuss the related works in a) semantic mapping, b) object detection and pose estimation, c) object contextual relations, and d) object temporal dynamics modeling.
  
\paragraph{Semantic Mapping}

Considering the plethora of work~\cite{kostavelis2015semantic} in the field of semantic mapping which vary in semantic representations, we limit our focus to the works that provide object-level semantics. Works in semantic SLAM ~\cite{bao2012semantic, salas2013slam++, bowman2017probabilistic} demonstrated SLAM at the object level. Similarly, we aim at providing a semantic map of the world at the object level, and we focus on mapping while making use of existing metric slam method (e.g., ORB-SLAM~\cite{mur2017orb}) to stay localized.

A widely used approach for semantic mapping is to augment 3D reconstructed map with objects.
Civera et al.~\cite{civera2011towards} ran an object detection thread parallelly with a monocular SLAM thread. They registered objects to the map by aligning the object faces relying on the SURF features. 
Ekvall et al.~\cite{ekvall2006integrating} actively recognized objects based on SIFT features, and integrated object recognition with SLAM for triangulation of object locations. But Civera et al. and Ekvall et al. did not deal with false detections, and their experiments were carried out in environments with no clutter.

To be robust to false detections, Pillai et al.~\cite{pillai2015monocular} proposed aggregating object evidence over multiple frames to get better detection, compared to single frame object detection. But their method relied on 3D geometric segmentation that singulates objects from the background, which is vulnerable when dealing with clutter. S\"{u}nderhauf et al.~\cite{sunderhauf2017meaningful} combined object detection over multiple frames and 3D geometric segmentation to get reasonable object boundaries. They produced 3D reconstructed map with object instance segments as central semantic entities. But their method did not provide object pose information, which is critical for robotic manipulation tasks.

Other works have focused on scene labeling of 3D map as a parallel SLAM thread is running in the background. 
People have proposed different methods for single frame scene labeling~\cite{zhao2014semantic, stuckler2015dense, mccormac2017semanticfusion, xiang2017rnn}, and fused labels across multiple frames to generate a dense 3D semantic map. Our work focus on detecting and localizing object entities in the environment, instead of dense labeling of every surfel or voxel in the reconstructed 3D map.

\paragraph{Object Detection and Pose Estimation}
Deep neural network based object detectors~\cite{redmon2016you, liu2016ssd, ren2017faster} are nowadays widely adopted for focusing attention in region of interest given an image. Works in object pose estimation adopt these object detectors to get prior on object locations. Zhen et al.~\cite{zeng2018srp} generated scene hypotheses based on object detections returned by R-CNN~\cite{girshick2014rich}, and they used Bayesian based bootstrap filter to estimate object poses. Similarly, Sui et al.~\cite{Suietal2017ijrr} and Narayanan et al.~\cite{Narayanan-2016-5537} proposed generative approach for object pose estimation given RGB-D observation. Discriminative object pose estimation methods use local~\cite{johnson1999using, rusu2009fast} or global~\cite{rusu2010fast, aldoma2012our} descriptors to estimate object poses via feature matching. However, feature-based methods are sensitive to the clutterness in the environment. Our work takes the generative approach and builds on Zhen et al.~\cite{zeng2018srp} for object pose estimation through Bayesian filtering, while~\cite{zeng2018srp} modeled objects independently and took single image at input, we model the contextual dependencies between objects and temporal consistency of each object instance given streaming data.

Works that simultaneously detect and localize objects are highly related to our work. Xiang et al.~\cite{xiang2017posecnn} proposed PoseCNN as a novel network for object detection and 6 DOF object pose estimation given a RGB image. Tremblay et al.~\cite{tremblay2018synthetically} and Tekin et al.~\cite{tekin2018real} converted the problem of simultaneous object detection and pose estimation into a problem of detecting the vertices of object bounding cuboid. Unlike these works that take single image as input and outputs deterministic estimate of object poses, our work maintains a belief over object classes and poses across observations.

Given streaming data, Salas-Moreno et al.~\cite{salas2013slam++} assumed repeated object instances in the environment to effectively recognize and localize objects, but their model lacks inter-object dependences. Tateno et al.~\cite{tateno20162} incrementally segmented 3D surface reconstructed by an underlying SLAM thread, then 3D segments were recognized as objects and object poses were estimated via 3D descriptor matching. Their work is similar to our work in terms of the output, but they depend on 3D geometric segmentation which is not guaranteed to segment objects out in clutter. In addition, they require dense SLAM with small voxel size which is hard to scale. 

\paragraph{Object Contextual Relations}
Contextual relations play a key role in modeling spatial relations between objects for scene understanding. Koppula et al.~\cite{koppula2011semantic} showed semantic labeling on point clouds using co-occurrence and geometric relations between objects. Jiang et al.~\cite{jiang2012learning} explored indirectly modeling object contextual relations by hallucinating human interactions with the environment. Similarly, ~\cite{galindo2005multi, heitz2008learning, kollar2009utilizing, espinace2013indoor, aydemir2013active} have proven modeling \textit{object-object} and \textit{object-place} contextual relations to be useful in place recognition, object detection and object search tasks.
In our work, we mainly utilize \textit{object-object} contextual relations in terms of co-occurrence and geometric relations. 


\paragraph{Object Temporal Dynamics Modeling} 
We need to maintain the belief over object poses even when objects are not being observed. 
Different types of the objects share different characteristics of dynamics. For example, structural objects such as furnitures tend to stay approximately at the same location, while small objects such as food items can often be moved from one place to another. Bore et al.~\cite{bore2018multiple} proposed to learn long-term object dynamics over multiple visits of the same environment. Russel et. al.~\cite{toris2017temporal} proposed a temporal persistence model to predict the probability of an object staying at the location where it is last observed after certain time period. We are inspired by the temporal persistence model proposed in~\cite{toris2017temporal}, and we reason about the possible locations of an object observed in the past based on the contextual relations between objects. 

\section{Problem Formulation}
We focus on semantic mapping at the object level. Our proposed {\em CT-Map} method maintains a belief over object classes and poses across an observed scene. We assume that the robot stays localized in the environment through an external localization routine (e.g., ORB-SLAM~\cite{mur2017orb}). The semantic map is composed by a set of $N$ objects $O=\{o^1, o^2, \cdots, o^N\}$. Each object $o^i=\{o^c, o^g, o^\psi\}$ contains the object class $o^c \in \mathcal{C}$, object geometry $o^g$, and object pose $o^\psi$, where $\mathcal{C}$ is the set of object classes $\mathcal{C}=\{c_1, c_2, \cdots, c_n\}$.

At time $t$, the robot is localized at $x_{t}$. The robot observes $z_t=\{I_t, S_t\}$, where $I_t$ is the observed RGB-D image, and $S_t$ are semantic measurements. The semantic measurements $s_k=\{s^s_k, s^b_k\} \in S_t$ are returned by an object detector (as explained in section \ref{sec:faster-rcnn}), which contains: 1) a object detection score vector $s^s_k$, with each element in $s^s_k$ denoting the detection confidence of each object class, and 2) a 2D bounding box $s^b_k$.

\begin{figure}[t]
\centering
\includegraphics[width=0.8\textwidth]{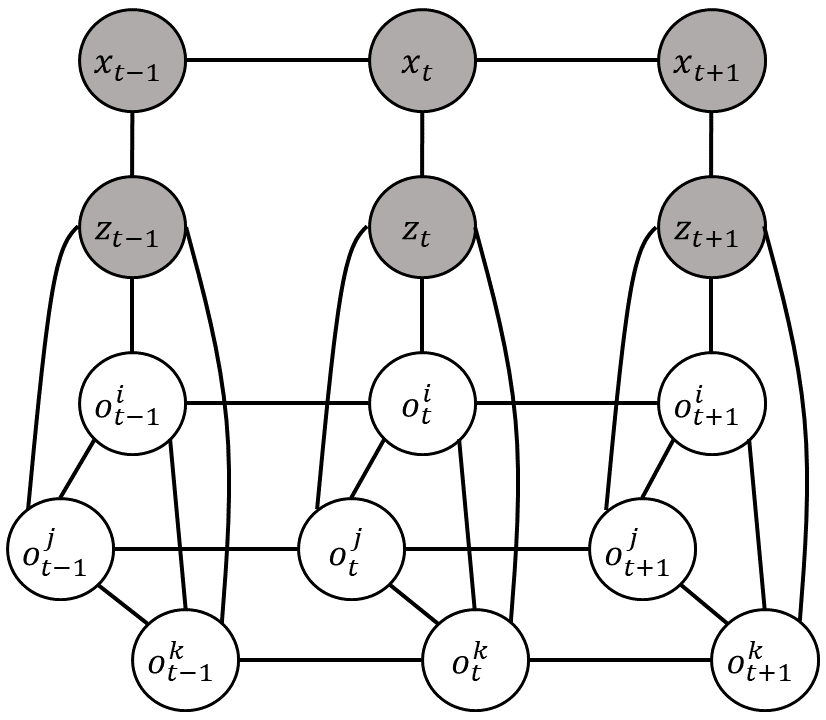}
\caption{Graphical model of the semantic mapping problem. Observed variables are robot poses $x_{t}$ and observations $z_t$. Unknown variables are objects $\{o^1, o^2, \cdots, o^N\}$. We compute the posterior over objects while modeling contexual relations between all pairs of objects at each time point, and temporal consistency of each object across consecutive time points.}\label{fig:graphical_model}
\end{figure}

We probabilistically formalize the semantic mapping problem in the form of a CRF, as shown in Figure \ref{fig:graphical_model}. Robot pose $x_{t}$ and observation $z_t$ are known. The set of objects $O$ are unknown variables. We model the contextual dependencies between objects and the temporal consistency of each individual object over time. The posterior probability of the semantic map is expressed as:
\begin{align}\label{eq:posterior_prob}
p(& O_{0:T} | x_{0:T},z_{0:T})= \nonumber \\
&\frac{1}{Z} \prod_{t=0}^T \prod_{i=1}^N \phi_p(o^i_t, o^i_{t-1}, u^i_{t-1}) \phi_m(o^i_t, x_t, z_t) \prod_{i,j} \phi_c(o_t^i, o_t^j)
\end{align}
where $Z$ is a normalization constant, and action applied to object $o^i$ at time $t$ is denoted by $u^i_t$. $\phi_p$ is the \textit{prediction potential} that models the temporal consistency of the object poses. $\phi_m$ is the \textit{measurement potential} that accounts for the observation model given 3D mesh of objects. $\phi_c$ is the \textit{context potential} that captures the contextual relations between objects.

\subsection{Prediction Potential}\label{sec:prediction_potential}
We use two different prediction models for predicting object pose, depending on whether the object is in the field of view or not. If the object is being observed, we model the action $u$ as a continuous random variable that follows a Gaussian distribution with zero mean and small variance $\Sigma$. This assumption leads to prediction of small object movements in 3D to be modeled as:
\[ o^\psi_t \sim \mathcal{N}(o^\psi_{t-1}, \, \Sigma) \]
which allows us to express the prediction potential as:
\begin{equation}\label{eq:prediction_potential}
\phi_p(o^i_t, o^i_{t-1}, u^i_{t-1}) = \exp(-(o^\psi_t-o^\psi_{t-1})^T \Sigma^{-1} (o^\psi_t-o^\psi_{t-1}))
\end{equation}

When object $o^i$ is not in the field of view for a significant period of time, it can be either located at the same location or moved to a different location due to the actions applied by other agents. As stated by Toris et al.~\cite{toris2017temporal}, the probability of the object $o^i$ still being at the same location where it was last seen is a function of time. To take into account the fact that object $o^i$ can be moved to other locations, we model the temporal action $u^i$ with a discrete random variable $\{u_{stay},\ u_{move}\}$. Specifically, $u_{stay}$ denotes no action and the object stays at the same location, and $u_{move}$ denotes a move action is applied and the object is moved to other locations. And these high-level actions follow certain distribution $p(u^i, \Delta t)$,
\begin{equation}\label{eq:temporal_action}
p(u^i = u_{stay}, \Delta t) = r_1 + r_2\exp(-\frac{\Delta t}{\mu^i})
\end{equation}
\begin{equation}
p(u_{stay}, \Delta t) + p(u_{move}, \Delta t) = 1
\end{equation}
where $r_1,\ r_2$ are constants, and $\Delta t$ is the time duration that object $o^i$ is not being observed. As $\Delta t$ increases, the probability of $u_{stay}$ decays, and eventually $p(u_{stay}, \Delta t) = r_1$ as $\Delta t \to \infty$. For different objects $o^i$, the coefficients $\mu^i$ that control the speed of the decay are different. We provide heuristic $\mu^i$ for different objects in our experiments, while these coefficients can also be learned as introduced by Toris et al.~\cite{toris2017temporal}.

\subsection{Measurement Potential}
The measurement potential of object $o^i_t$ is expressed as:
\[ 
\phi_m(o^i_t, x_t, z_t)=\begin{cases}
               \delta, & \text{if}\ o^i_t \ \text{is out of view} \\
               g(o^i_t, x_t, z_t), & \text{otherwise}
            \end{cases}
\]

We use non-zero constant $\delta$ to account for cases where objects are not in the field of view. $g(o^i_t, x_t, z_t)$ measures the compatibility between the observation $z_t$ and $o^i_t,\ x_t$,
\[g(o^i_t, x_t, z_t) = \sum_{s_k \in S_t} h(o^i_t, s^s_k) l(s^b_k, b(o^i_t, x_t)) f(o^i_t, x_t, I_t) \]
where $h(o^i_t, s^s_k)$ is the confidence score of class $o^{c}_t$ from the detection confidence vector $s^s_k$. Function $l$ evaluates the intersection over minimum area of two bounding boxes. $b(o^i_t,x_t)$ is the minimum enclosing bounding box of projected $o^i_t$ in image space based on $x_t$.

We assume known 3D mesh models of objects. Function $f(o^i_t, x_t, I_t)$ computes the similarity between the projected $o^i_t$ and $I_t$ inside bounding box $b(o^i_t,x_t)$, as explained in detail in section \ref{sec:impl-f}. In the case that robot has observed object $o^i$ in the past, and the belief over $o^i$ indicates that it is in the field of current view of the robot. If the robot cannot detect object $o^i$, then the object could be occluded, in which case we use $g(o^i_t, x_t, z_t)=f(o^i_t, x_t, I_t)$ for the object to be potentially localized.

\subsection{Context Potential}
There exist common contextual relations between object categories across all environments. For example, a cup would appear on a table much more often than on the floor, and a mouse would appear besides a keyboard much more often than besides a coffee machine. We refer to these common contextual relations as \textit{category-level} contextual relations. In a specific environment, there exist contextual relations between certain object instances. For example, a TV always stays on a certain table, and a cereal box is usually stored in a particular cabinet. We refer to these contextual relations in a specific environment as \textit{instance-level} contextual relations.

We manually encode \textit{category-level} contextual relations as prior knowledge to our model, which also can be learned from public scene dataset (e.g., McCormac et al.~\cite{mccormac2016scenenet}). Because \textit{instance-level} contextual relations vary across different environments, these relations of a specific environment must be learned over time. The \textit{context potential} is composed by \textit{category-level} potential $\phi_{cat}$ and \textit{instance-level} potential $\phi_{ins}$,
\begin{equation}\label{eq:contextual-potential}
\phi_c(o^i_t, o^j_t) = w_1 \phi_{cat}(o^i_t, o^j_t) + w_2 \phi_{ins}(o^i_t, o^j_t)
\end{equation}
We model $\phi_c(o^i_t, o^j_t)$ as mixture of Gaussians, with $\phi_{cat}(o^i_t, o^j_t)$ and $\phi_{ins}(o^i_t, o^j_t)$ each being a Gaussian component.

In our experiments, we manually designed $\phi_{cat}$ as prior knowledge, and $\phi_{ins}$ is updated via Bayesian updates. The principle  while designing $\phi_{cat}$ follows two constraints: 1) simple physical constraints such as no object intersection is allowed, and objects should not be floating in the air, and 2) object pairs that belong to the same category co-occur more often than objects from different categories.

\section{Inference}
\begin{algorithm}[t!]
\KwIn{Observation $z_t$, robot pose $x_t$, particle set for each object $Q_{t-1}^i=\{\langle o^{i(k)}_{t-1}, \alpha^{i(k)}_{t-1} \rangle | k=1, \cdots, M \}$}
Resample $M$ particles $o^{i(k)}_{t-1}$ from $Q_{t-1}^i$ with probability proportional to importance weights $\alpha^{i(k)}_{t-1}$ \;
\For{$i=1, \cdots, N$}{
	\For{$k=1, \cdots, M$}{
		Sample $o^{i(k)}_t \sim \phi_p(o^{i}_t, o^{i(k)}_{t-1}, u_{t-1})$ \;
		Assign weight $\alpha^{i(k)}_t \propto \phi_m(o^{i(k)}_t, x_t, z_t) \prod_{j \in \Gamma(i)} \phi_c(o^{i(k)}_t, o^j_{t-1})$ \;
	}
}
 \caption{Particle filtering in {\em CT-Map}} \label{alg:particle_filter}
\end{algorithm}

We propose a particle filtering based algorithm to perform inference in {\em CT-MAP}, as given in Algorithm \ref{alg:particle_filter}. Nonparametric Belief Propagation~\cite{sudderth2003nonparametric}~\cite{isard2003pampas} is not directly applicable to our problem because we are dealing with high-dimensional data. Sener et al. proposed recursive CRF~\cite{Sener-RSS-15} that deals with discrete hidden state with forward-backward algorithm, while our hidden state is mixed, i.e., object class label in discrete space and object pose in continuous space. 

Instead of estimating the posterior of the complete history of objects $O_{1:T}$ as expressed in Equation \ref{eq:posterior_prob}, {\em CT-Map} can recursively estimate the posterior of each object $o^i_t \in O_t$. This approach to inference is similar to the CRF-filter proposed by Limketkai et al.~\cite{limketkai2007crf}. We represent the posterior of object $o^i_t$ with a set of $M$ weighted particles, i.e., $Q_t^i=\{\langle o^{i(k)}_t, \alpha^{i(k)}_t \rangle | k=1, \cdots, M \}$, where $o^{i(k)}_t$ contains object class and pose information as introduced in \ref{sec:prediction_potential}, and $\alpha^{i(k)}_t$ is the associated weight for the $k^{th}$ particle. In each particle filtering iteration, particles are first resampled based on their associated weights, then propagated forward in time through object temporal consistency, and re-weighted according to the measurement and context potentials.

We associate bounding boxes across consecutive frames based on their overlap. Only if a bounding box has been consistently associated for certain number of frames will we start initiating object class and pose estimation for that bounding box. The initial set of particles given a detected bounding box $s^b_k$ are drawn as following: 1) first we sample the object class $o^c$ based on the corresponding detection confidence score vector $s^s_k$; 2) then we sample the 6 DOF object pose $o^\psi$ inside $s^b_k$, by putting the object center around the 3D points at the center region of $s^b_k$, with orientation uniformly sampled.

To sample the pose of $o^{i(k)}_t$ from $\phi_p(o^i_t, o^{i(k)}_{t-1}, u_{t-1})$ (Step 4 in Algorithm \ref{alg:particle_filter}), there are two cases as following: 
\begin{itemize}
\item
If $o^{i(k)}_{t-1}$ is within the field of view of the robot, we sample $o^{i(k)}_t$ according to Equation \ref{eq:prediction_potential}.
\item
If $o^{i(k)}_{t-1}$ is not within the field of view of the robot, we first sample the high-level action $\{u_{stay}, u_{move}\}$ according to Equation \ref{eq:temporal_action}. 
\begin{itemize}
\item
If $u_{stay}$ is sampled, then $o^{i(k)}_t$ is sampled based on Equation \ref{eq:prediction_potential}. 
\item
If $u_{move}$ is sampled, then another object $o^j$ is uniformly sampled from $O \setminus o_i$, which indicates the place that $o^i$ has been moved to. $o^{i(k)}_t$ is then sampled from the region that $o^j$ can physically support.
\end{itemize}
\end{itemize}

In step 5 of Algorithm \ref{alg:particle_filter}, we use $\Gamma(i)$ to denote the indices of objects that are in the neighborhood of object $o^{i(k)}_t$. Because each neighbor object $o^j_{t-1}$ is represented by $M$ particles, it is computationally expensive to evaluate the context potential $\phi_c(o^{i(k)}_t, o^j_{t-1})$ against each particle of $o^j_{t-1}$. Thus, we only evaluate the context potential against the most likely particle of $o^j_{t-1}$.


\section{Implementation}
\subsection{Faster R-CNN object detector}\label{sec:faster-rcnn}
We deploy Faster R-CNN~\cite{ren2017faster} as our object detector. Given the RGB channel of our RGB-D observation, we apply the object detector and get the bounding boxes from the region proposal network, along with the corresponding class score vector. Then we apply non-maximum suppression to these boxes and merge boxes that have Intersection Over Union (IoU) larger than 0.5.
For training, our dataset has 970 groundtruth images for 13 object classes. Each image has around $10$ labeled objects. We fine-tuned the object detector based on VGG16~\cite{DBLP:journals/corr/SimonyanZ14a} pretrained on COCO~\cite{coco}. In case of overfitting, we fine-tuned the network for $3000$ interations with $0.001$ learning rate.

\subsection{Similarity function $f(o^i_t, x_t, I_t)$}\label{sec:impl-f}
We assume as given the 3D mesh model of objects. Thus, we can render the depth image of $o^i_t$ based on its object class $o^c$ and 6 DOF pose $o^\psi$ in the frame of $x_t$. With rendered depth image $I(o^i_t, x_t)$:
\begin{equation} \label{eq:similarity}
f(o^i_t, x_t, I_t) = e^{-\lambda d(I(o^i_t, x_t), I_t)}
\end{equation}
where $\lambda$ is a constant scaling factor. $d(I(o^i_t, x_t), I_t)$ is the sum of squared differences between the depth values in observed and rendered depth images.

\section{Experiments}
We collected our indoor scene dataset with a Michigan Progress Fetch robot for evaluation on our proposed {\em CT-Map} method. Our indoor scene dataset contains 20 RGB-D sequences of various indoor scenes. We measure the quality of inference for various scenes in terms of 1) object detection and 2) pose estimation. Thus, we follow the mean average precision (mAP) metric and 6 DOF pose estimation accuracy for benchmarking our method. We also show qualitative examples of our semantic maps in Figure \ref{fig:teaser}. More qualitative examples are provided in the video\footnote{\url{https://youtu.be/W-6ViSlrrZg}}.

Across all experiments, we use $w_1=w_2=0.5$ in Equation \ref{eq:contextual-potential} to treat \textit{category-level} and \textit{instance-level} potentials equally. If an object has not been observed for infinite long period of time, we assume that object has equal probabilities of either staying at the same location or not. Thus, we use $r_1=r_2=0.5$ in Equation \ref{eq:temporal_action}. 

\begin{table}[t!]
\centering
\begin{tabular}{|c|c|c|c|} 
 \hline 
   & Faster R-CNN & {\em T-Map} & {\em CT-Map} \\ [0.0ex] 
 \hline
 mAP & 0.607 & 0.715 & 0.871 \\ [0ex] 
 \hline
\end{tabular}
\caption{mAP on our scene dataset.}
\label{table:meanAP}
\end{table}

\subsection{Object Detection}
We have noisy object detections coming from baseline Faster R-CNN object detector, while {\em CT-Map} can correct some false detections by modeling the object class as part of our hidden state. To evaluate the object detection performance of {\em CT-Map}, we take the estimated 6 DOF pose of all objects in the scene at the end of each RGB-D sequence in our dataset, and project them back onto each camera frame in that sequence to generate bounding boxes with class labels. 
We run two semantic mapping processes by considering different sets of potentials: 1) Temporal Mapping ({\em T-Map}): we consider prediction potential in the CRF model; 2) Contextual Temporal Mapping ({\em CT-Map}): we consider both prediction and context potential in the CRF model, which is the proposed method. For both {\em T-Map} and {\em CT-Map}, we include the measurement potential on observation.

We use mAP as our object detection metric. As shown in Table \ref{table:meanAP}, {\em T-Map} improves upon the baseline method Faster R-CNN by incorporating prediction and observation potentials, and {\em CT-Map} improves the performance further by additionally incorporating context potential. Faster R-CNN did not perform quite well on the test scenarios because the training data do not necessarily cover the variances encountered at test time.
Though the performance of Faster-RCNN can be further improved by providing more training data, {\em CT-Map} provides more robust object detection when training remains limited.

\begin{figure*}[t!]
    \centering
   	\includegraphics[width=\textwidth]{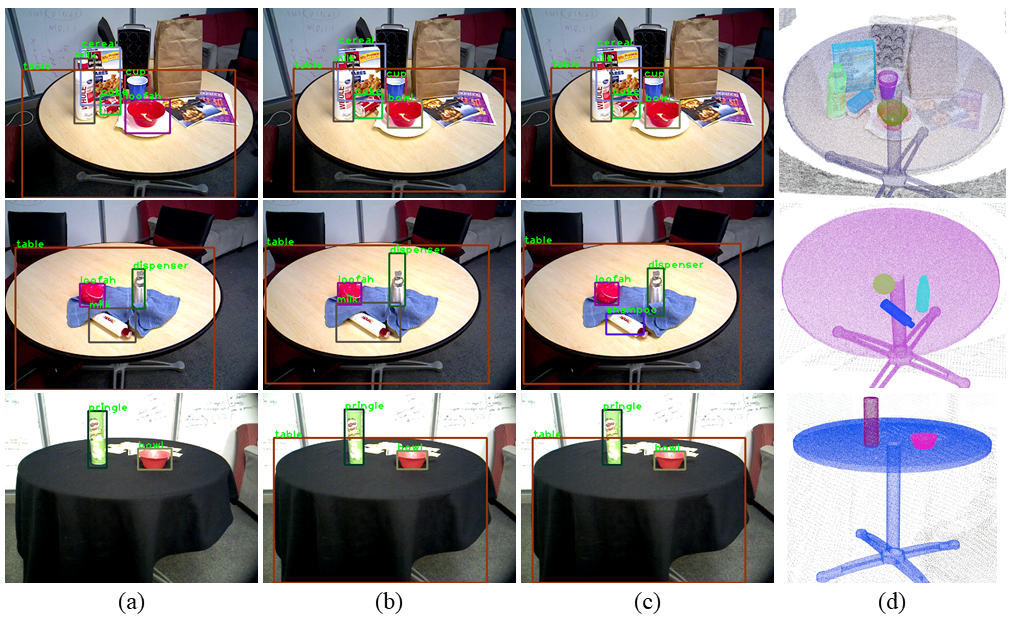}
 \caption{Mapping examples highlighting detection improvements: (a) raw detection results from baseline Faster R-CNN; (b) detection results from {\em T-Map} when only considering measurement and prediction potential; (c) detection results {\em CT-Map} when considering measurement, prediction and context potential; (d) 6 DOF object pose estimates from {\em CT-Map}. We generate bounding boxes in column (b) and (c) by projecting the localized 3D objects into 2D image space, and finding the minimum enclosing boxes of the projections. The first row shows Faster R-CNN gives false detection on the red bowl as "loofah", while both {\em T-Map} and {\em CT-Map} correct the wrong label "loofah" into "bowl". The second row shows Faster R-CNN gives false detection on the shampoo bottle as "milk", and {\em T-Map} fails to correct the wrong label because the geometry of milk and shampoo is similar, while {\em CT-Map} successfully corrects the wrong label into "shampoo" based on the context. The third row shows Faster R-CNN does not detect the table due to the appearance change induced by the table cloth, while both {\em T-Map} and {\em CT-Map} successfully detect and localize the table. Because the table used to be observed around that location in the past, and our methods benefit from modeling the temporal consistency of object poses. (Best viewed in color)}
 \label{fig:detection_highlight}
\end{figure*}

In some cases, objects are not being reliably detected by Faster R-CNN due to occlusion. If an object has been observed in the environment in the past, our method makes predictions on locations that objects can go by modeling the temporal consistency of objects. Thus, even if a detection is not fired on the object due to occlusion, our method can still localize the object and claim a detection. However, in cases where an object is severely occluded and the depth observation lacks enough geometric information from the object, our method will not be able to localize the object. Example detection results highlighting the benefits of the proposed method compared to baseline Faster R-CNN are shown in Figure \ref{fig:detection_highlight}.


\subsection{Pose Estimation}
For each RGB-D sequence in our dataset, we locate the frames that each object is last seen, and project the depth frame back into 3D point clouds using known camera matrix. We then manually label the ground truth 6 DOF pose of objects. We compare the estimated object poses at the end of each RGB-D sequence against the ground truth. 

Pose estimation accuracy is measured as $accuracy = \frac{N_{correct}}{N_{total}}$, 
where $N_{correct}$ is the number of objects that are considered correctly localized, and $N_{total}$ is the total number of objects that are present in the dataset. If the object pose estimation error falls under certain position error threshold $\Delta t$ and rotation error threshold $\Delta \theta$, we claim that the object is correctly localized. $\Delta t$ is the translation error in Euclidean distance, and $\Delta \theta$ is the absolute angle difference in orientation. For symmetrical objects, the rotation error with respect to the symmetric axis is ignored.

We apply the Iterative Closest Point (ICP) and Fast Point Feature Histogram (FPFH)~\cite{rusu2009fast} algorithms as our baselines for 6 DOF object pose estimation. For each RGB-D sequence in our dataset, we take the 3D point clouds of the labeled frame, and crop them based on ground truth bounding boxes. These cropped point clouds are given to the baselines as observations, along with object 3D mesh models. ICP and FPFH are applied to register the object model to the cropped observed point cloud. We allow maximum iterations of 50000.

Our proposed method {\em CT-Map} significantly outperforms ICP and FPFH by a large margin. As our generative inference iteratively samples object pose hypotheses and evaluates them against the observations, {\em CT-Map} does not suffer from local minima as much as discriminative methods such as ICP and FPFH.

\begin{figure}
\centering
\includegraphics[width=1\textwidth]{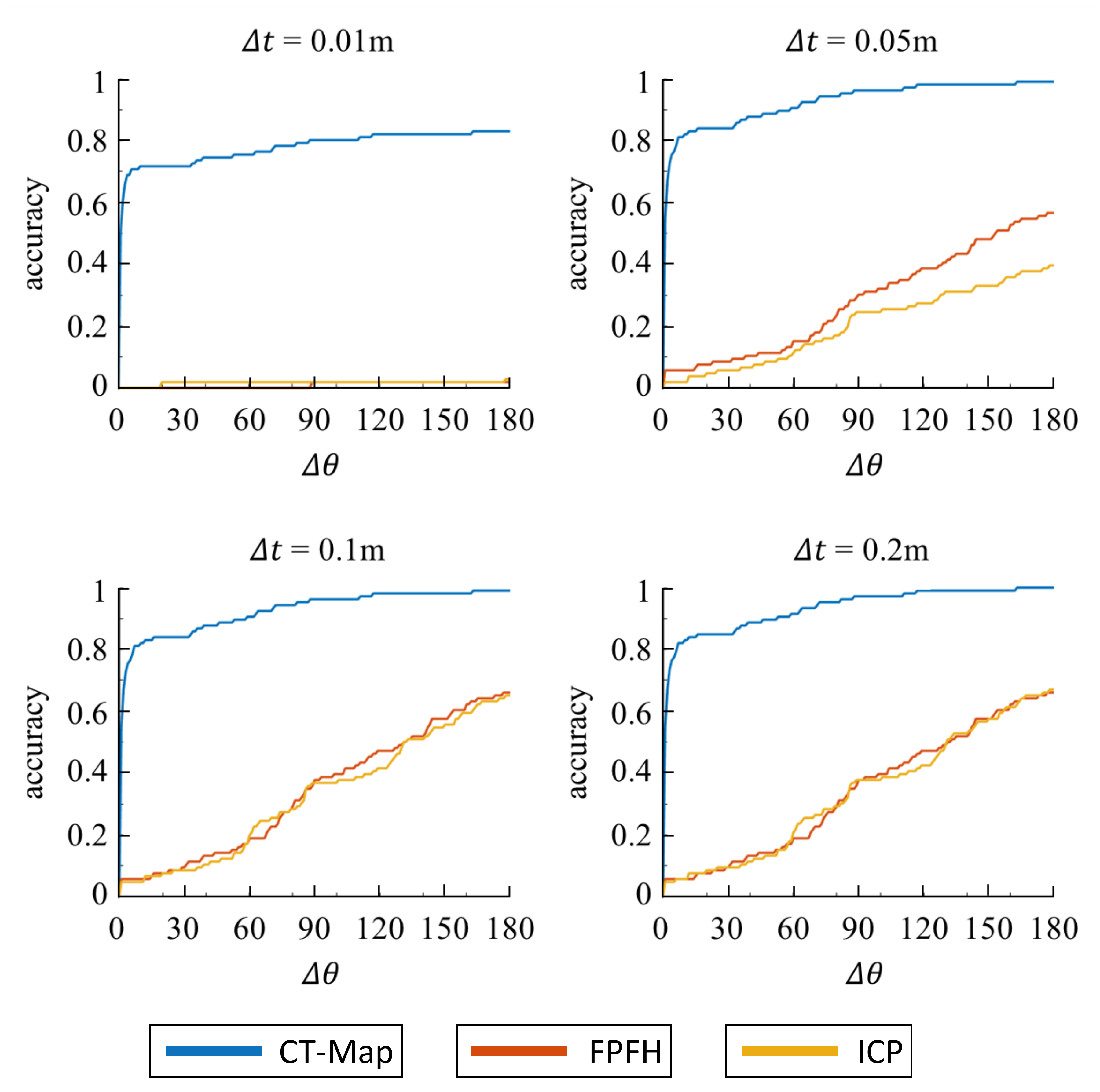}
\caption{Object pose estimation of {\em CT-Map}, compared with FPFH and ICP based baselines. Different plots correspond to different pose estimation correctness criteria defined by position error threshold $\Delta t$ and rotation error threshold $\Delta \theta$. Our method outperforms FPFH and ICP with a large margin.}
\label{fig:benchmark_all_delta_}
\end{figure}

\section{Conclusion}


We propose a semantic mapping method {\em CT-Map} that simultaneously detects objects and localizes their 6 DOF pose given streaming RGB-D observations. {\em CT-Map} represents the semantic map with a belief over object classes and poses. We probabilistically formalize the semantic mapping problem in the form of a CRF, which accounts for contextual relations between objects and temporal consistency of object poses, as well as measurement potential on observation. We demonstrate that {\em CT-Map} outperforms Faster R-CNN in object detection and FPFH, ICP in object pose estimation. In the future, we would like to investigate the inference problem of object semantic locations given partial observations of an environment, e.g., inferring a query object to be on a dining table, or in a kitchen cabinet. Ideally, maintaining a belief over object semantic locations can serve as a notion of generalized object permanence, and facilitate object search tasks.

\section*{Acknowledgement}
This work was supported in part by NSF award IIS-1638060.
\balance
\bibliographystyle{abbrv}
\bibliography{my_bib}

\end{document}